\documentclass[twocolumn,letterpaper,10pt]{IEEEAerospaceCLS}

\usepackage[utf8]{inputenc}
\usepackage{amssymb, amsfonts, amstext, bm, mathtools} %
\usepackage[pdftex]{graphicx} %
\graphicspath{{figs/}}

\usepackage{cite}

\normalsize
\usepackage{tabularray}
\SetTblrInner[tblr]{rowsep=0pt} %
\newcommand{\TCategoryAboveSpace}{\rule{0pt}{15pt}}
\UseTblrLibrary{booktabs, siunitx}

\usepackage[linesnumbered,ruled]{algorithm2e}
\usepackage{setspace}
\usepackage{algpseudocode}

\SetCommentSty{mycommfont}

\usepackage{enumitem}
\setlist[enumerate]{itemindent=0em, leftmargin=1.5em, itemsep=0.1em}
\setlist[itemize]{itemindent=0em, leftmargin=1.5em, itemsep=0.1em}

\usepackage[hidelinks]{hyperref}
\usepackage[capitalise]{cleveref}

\crefname{problem}{Problem}{Problems}

\usepackage[font=bf]%
            {caption}
\usepackage{subcaption}

\newtheorem{theorem}{Theorem}[section]
\newtheorem{problem}[theorem]{Problem}

\SetKwInput{KwInput}{Input}
\SetKwInput{KwOutput}{Output}
\newcommand{\nonl}{\renewcommand{\nl}{\let\nl\oldnl}}%
\newcommand{\nosemic}{\renewcommand{\@endalgocfline}{\relax}}%
\newcommand{\dosemic}{\renewcommand{\@endalgocfline}{\algocf@endline}}%

\usepackage{xcolor}

\newcommand{\mcal}[1]{{\mathcal{#1}}}
\newcommand{\mbb}[1]{{\mathbb{#1}}}

\title{Safe Mission-Level Path Planning for Exploration of Lunar Shadowed Regions by a Solar-Powered Rover}

\author{
Olivier Lamarre* \\
STARS Laboratory \\
Institute for Aerospace Studies\\
University of Toronto\\
4925 Dufferin St. \\
Toronto, ON M3H\hspace{0.25em}5T6
\and
Shantanu Malhotra** \\
Jet Propulsion Laboratory \\
California Institute of Technology\\
4800 Oak Grove Dr. \\
Pasadena, CA 91109 \\ \\
*\{first.last\}@robotics.utias.utoronto.ca \\
**shantanu.malhotra@jpl.nasa.gov
\and
Jonathan Kelly* \\
STARS Laboratory \\
Institute for Aerospace Studies\\
University of Toronto\\
4925 Dufferin St. \\
Toronto, ON M3H\hspace{0.25em}5T6
\thanks{\footnotesize 979-8-3503-0462-6/24/$\$31.00$ \copyright2024 IEEE}
}

\begin{document}
\maketitle

\thispagestyle{plain}
\pagestyle{plain}

\begin{abstract}
Exploration of the lunar south pole with a solar-powered rover is challenging due to the highly dynamic solar illumination conditions and the presence of permanently shadowed regions (PSRs). In turn, careful planning in space and time is essential.
Mission-level path planning is a global, spatiotemporal paradigm that addresses this challenge, taking into account rover resources and mission requirements.
However, existing approaches do not proactively account for random disturbances, such as recurring faults, that may temporarily delay rover traverse progress.
In this paper, we formulate a chance-constrained mission-level planning problem for the exploration of PSRs by a solar-powered rover affected by random faults.
The objective is to find a policy that visits as many waypoints of scientific interest as possible while respecting an upper bound on the probability of mission failure.

Our approach assumes that faults occur randomly, but at a known, constant average rate.
Each fault is resolved within a fixed time, simulating the recovery period of an autonomous system or the time required for a team of human operators to intervene.
Unlike solutions based upon dynamic programming alone, our method breaks the chance-constrained optimization problem into smaller offline and online subtasks to make the problem computationally tractable.
Specifically, our solution combines existing mission-level path planning techniques with a stochastic reachability analysis component.
We find mission plans that remain within reach of safety throughout large state spaces.
To empirically validate our algorithm, we simulate mission scenarios using orbital terrain and illumination maps of Cabeus Crater.
Results from simulations of multi-day, long-range drives in the LCROSS impact region are also presented.
\vspace{-2\baselineskip}
\end{abstract}

\tableofcontents

\addtocounter{footnote}{-1}

\section{Introduction}

Over the past two decades, lunar probes have detected large quantities of water and other volatiles near the surface of permanently shadowed regions (PSRs)~\cite{colaprete_detection_2010,li_direct_2018}.
These observations have sparked international efforts towards developing robotic missions to further characterize the distribution of these resources for scientific and human exploration purposes.
Numerous recent concepts requiring long-range mobility have been formulated~\cite{flahaut_regions_2020, lemelin_framework_2021, losekamm_assessing_2022, mazarico_sunlit_2023}.
The solar-powered Volatiles Investigating Polar Exploration Rover (VIPER), one of the first missions of this kind, is expected to explore lunar PSRs near Nobile Crater in late 2024 or early 2025~\cite{nasa_viper_2020}.

A primary challenge of long-range mobility with a solar-powered rover at the lunar south pole is the highly dynamic illumination caused by a low Sun elevation above the horizon.
Global planning in space and time is required to ensure sufficient solar exposure to prevent battery depletion.
In a scientific exploration context, the ability of a solar-powered rover to investigate distant regions of interest strongly depends on \textit{when} and \textit{where} the rover navigates during its mission.
Mission-level path planning is a methodology that addresses this challenge.
A mission-level planner finds admissible spatiotemporal trajectories while taking into account high-level mission objectives and resource constraints~\cite{tompkins_mission-level_2006}.

Existing mission-level path planners generally rely on offline optimization techniques or do not proactively account for stochastic disturbances, such as recurring faults, that may temporarily halt long-range traverses.
When exploring PSRs or any area severely deprived of sunlight, the lethality of a fault depends on a variety of factors including the location and time at which it occurs.
For example, a fault that occurs in a well-insolated area may not present any danger.
A similar fault occurring in the shade, however, might lead to the loss of the rover if the delay prevents a critical solar charging event from happening in the near future.
In the presence of stochastic disturbance, deterministic safety constraints cannot ensure rover safety.
For instance, there is always a non-zero, albeit small, probability of experiencing numerous faults inside of a PSR to the point of running out of energy, regardless of the rover's battery capacity.
Instead, a constraint on the \textit{probability} of mission failure (also called a chance-constraint) is more appropriate.

\begin{figure*}[t]
    \centering
    \includegraphics[width=\textwidth]{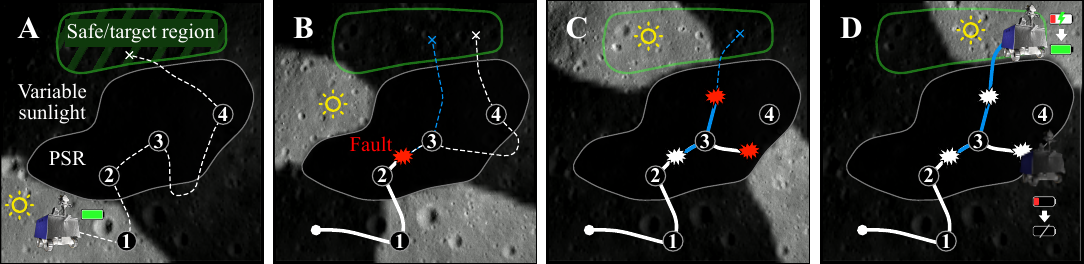}
    \caption{The importance of risk-aware planning when exploring PSRs with a solar-powered rover affected by recurring, random faults.
    Subfigure A shows a nominal mission plan visiting waypoints of interest in a given order.
    After a fault/delay occurs during the traverse (Subfigure B), risk-aware planning (blue dashed line) suggests an early PSR exit while risk-agnostic planning (white dashed line) finds an updated plan visiting the fourth waypoint, unaware of the danger of this traverse schedule.
    If a second fault occurs inside of the PSR (Subfigure C), a rover following the risk-agnostic plan misses a crucial solar charging period and battery energy depletion becomes unavoidable (Subfigure D).
    A rover following the risk-aware plan, on the other hand, is still capable of reaching the designated target region safely. Background image courtesy of NASA and Arizona State University. VIPER render courtesy of NASA.
    }
    \label{fig:overview}
    \vspace{-2mm}
\end{figure*}

In this work, we formulate and solve a joint chance-constrained mission-level path planning problem.
Given an ordered set of waypoints of (scientific) interest, we seek a policy (an online decision-making scheme) that visits as many as possible while respecting an upper bound on the probability of mission failure, or \textit{risk}.
A mission failure occurs when the rover terminates its traverse outside of a pre-determined safe region, beyond a specified time limit, or without sufficient energy to ensure long-term survival.
We assume that stochastic faults temporarily delay the traverse of a rover at a known average spatial rate (for example, one fault every five kilometres driven), similar to previous mission studies~\cite{robinson_intrepid_2020,keane_endurance_2022}.
In practice, fault-induced delays could represent the time required for an autonomous system to resume navigation, or for human operators to intervene.
The overall challenge is conceptually illustrated in \Cref{fig:overview}.

Our proposed algorithm combines conventional, offline mission-level path planning with an online safe stochastic reachability analysis component.
The former enables searches through large state spaces while the latter facilitates the calculation of a lower bound on risk for all candidate mission plans. 
As part of our framework, we devise a type of risk-bounded AND-OR (AO) planning method that generates \textit{partial} policy trees.
From a given start state, these trees indicate the best nominal plan and the corresponding execution risk by accounting for where and when faults may occur.
As the rover carries out its mission, these policy trees indicate how the rover should behave to reach more waypoints, or whether it should retreat to a safe state.
As such, our algorithm guides the rover towards regions of interest while %
ensuring that it is within reach (in a probabilistic sense) of a safe region, given a stochastic fault model.
The contributions of our paper are threefold:

\begin{enumerate}
    \item Formulation of a joint chance-constrained mission-level online planning problem for the exploration of environments deprived of solar illumination using solar-powered rovers.
    \item Creation of a risk-bounded AO search technique to construct partial policy trees across large state spaces, enabled by the combination of conventional mission-level path planning and stochastic reachability analysis.
    \item Validation of our approach through simulations of multi-kilometre scale traverses through lunar PSRs using terrain and solar illumination orbital maps of the Cabeus Crater region.
\end{enumerate}

The remainder of the paper is structured as follows.
\Cref{sec:relatedwork} summarizes relevant work in spatiotemporal and mission-level path planning for global navigation with solar-powered rovers.
We also provide an overview of chance-constrained optimization, focused on those frameworks previously applied to planetary exploration-related scenarios.
\Cref{sec:problemstatement} formally introduces the chance-constrained mission-level online planning problem.
Our proposed algorithm is detailed in \Cref{sec:method}.
In \Cref{sec:experiments}, we empirically compare our approach against solutions from another chance-constrained optimization solver by simulating drives at the lunar south pole using orbital terrain and illumination maps of Cabeus Crater.
Lastly, we simulate a large-scale mission scenario in the vicinity of the Lunar Crater Observation and Sensing Satellite (LCROSS) impact site in \Cref{sec:lcross}.%

\section{Related Work}
\label{sec:relatedwork}

We begin by reviewing approaches to spatiotemporal planning for solar-powered long-range surface mobility, most of which assume a deterministic rover-environment model.
Then, we identify relevant research in chance-constrained optimization and discuss preliminary applications to planetary exploration.

\subsection{Spatiotemporal and Mission-Level Global Path Planning}

Sun-synchronous planetary exploration, which involves maximizing sunlight exposure to enable long-term missions with solar-powered rovers, was an early motivation for spatial and temporal planning~\cite{whittaker_sun-synchronous_2000}.
Field experiments demonstrating long-range mobility with predictive energy management were first carried out in the Canadian Arctic using an early version of the Temporal Mission Planner for the Exploration of Shadowed Terrain (TEMPEST).
Unlike conventional mission schedulers, TEMPEST focuses on the relationship between resource management and route planning, a central challenge in sun-synchronous navigation~\cite{wettergreen_sun-synchronous_2005}.
Enabling a new global autonomy paradigm called mission-level path planning, TEMPEST is capable, under certain circumstances, of reactive plan repairs during traverse execution~\cite{tompkins_mission-level_2006}.
This optimization framework supported a series of long-range biological surveys in the Atacama desert in Chile \cite{wettergreen_second_2005, wettergreen_long-distance_2008}.

Other global spatiotemporal planning techniques have been developed for mobility at the lunar poles.
In~\cite{otten_planning_2015}, routes that are constantly illuminated are found via connected component analysis of solar illumination data across the spatiotemporal domain.
The same authors also account for Earth communication blackouts in~\cite{otten_strategic_2018}.
An offline mission-level path planner based on a forward-time A* search algorithm is presented in~\cite{cunningham_accelerating_2017} and introduces heuristics to accelerate the generation of long-range traverse plans through waypoints of interest.

Very few long-range mobility planning approaches for the lunar south pole proactively account for stochastic disturbances.
In~\cite{inoue_spatio-temporal_2021}, the probability of different numbers of faults is modelled using a binomial distribution, and this probability is incorporated into a spatiotemporal planner.
Trajectories entering shadowed areas are avoided.
The algorithms presented in~\cite{hu_planning_2022} heuristically combine safety functions to find feasible sun-synchronous routes.
In~\cite{lamarre_recovery_2023}, a definition of safety for solar-powered rovers exploring PSRs at the lunar south poles is provided and utilized as part of a stochastic reachability problem.
Safe recovery policies are identified from different start states for long-range traverses delayed by random faults.
This work, however, does not account for mission-level objectives such as visiting waypoints of scientific interest.
Safe mission-level path planning for exploring PSRs at the lunar south pole is an unresolved challenge.

\subsection{Chance-Constrained Optimization for Planetary Exploration}

Chance-constrained temporal optimization problems, such as risk-bounded event scheduling in simple temporal networks, have been studied in~\cite{fang_chance-constrained_2014,wang_chance-constrained_2015}.
Conditional Planning for Autonomy with Risk (CLARK) is a high-level framework for generating task plans under temporal uncertainty.
CLARK solves risk-bounded partially observable Markov decision processes (POMDPs) and has been demonstrated in planetary rover exploration problems~\cite{santana_risk-aware_2016,mcghan_resilient_2016}.
Internally, this framework employs the p-Sulu~\cite{ono_probabilistic_2013} and PARIS~\cite{santana_paris_2016} algorithms for risk-aware path planning and scheduling, respectively.
In this case, spatial and temporal planning occur separately.
Additionally, the p-Sulu algorithm alone is limited to simple environments, since the computation time increases exponentially with the number of obstacles in the environment.

Internally, CLARK relies on an AND/OR (AO)-based model that has been employed for decades in decision-making tasks where a finite number of outcomes are possible following an action~\cite{nilsson_principles_1982}.
Most AO search algorithms build acyclic hypergraphs with two different types of nodes.
An ``OR'' node represents a system state.
Successor nodes are of type ``AND'', representing all actions that can be taken from the parent OR node.
The children of each AND node are, in turn, new OR nodes representing all possible outcomes associated with the corresponding state transition.
The probabilities associated with such transition are encoded in the hyperedges connecting OR nodes to their predecessor/parent AND node.

The AO-based algorithm that CLARK relies on, named risk-bounded AO-star (RAO*), employs admissible heuristics to estimate future rewards and risks, and prunes exceedingly dangerous states from the solution tree~\cite{santana_rao_2016}.
Employing RAO* on large problem instances is computationally costly, however; implementations that use an iterative framework over a receding horizon~\cite{huang_hybrid_2018} and adaptations to fully observable problems with expensive action risk calculations~\cite{gutow_andor_2022} also exist.

Dynamic programming (DP) can be harnessed to find optimal policies subject to probabilistic constraints for planetary exploration-related tasks.
The authors of~\cite{kuwata_risk-constrained_2012} solve a multi-stage, chance-constrained decision-making problem for Martian entry descent and landing (EDL) and surface mobility planning.
In~\cite{ono_chance-constrained_2015}, a similar problem is addressed using an approximation algorithm and the corresponding optimality error bounds are derived.
Unfortunately, as with any DP-based algorithm, these approaches suffer from scalability issues because they must iterate over the entire state space.

\section{Problem Statement}
\label{sec:problemstatement}

We formulate an optimization problem where a solar-powered rover must visit as many waypoints of scientific interest as possible in a given order.
In this regard, our formulation draws inspiration from existing deterministic mission-level path planning problems introduced in \cite{tompkins_mission-level_2006, cunningham_accelerating_2017}.
However, we also require the rover to terminate its mission in a predetermined safe region of the state space.
Since the rover is temporarily halted by stochastic faults, we impose a threshold on the probability of mission failure (i.e., the probability that the rover does not reach safety, or \textit{risk}).
We begin by detailing the dynamics of our stochastic system and then formally introduce the corresponding chance-constrained (CC) mission-level online planning problem.

\subsection{Mobility Model Dynamics}

Consider a solar-powered rover moving on a grid $\mcal{C}$ (defined by an orbital map), in which $n_w$ cells represent locations of scientific interest.
Let $\mcal{W}$ be the list of grid cells (or \textit{waypoints}) in the order\footnote{During real rover missions, waypoint ordering is typically inferred from mission constraints or scientific domain knowledge.
Here, it also keeps the search problem tractable.} that they should be visited:
\begin{equation}
    \mcal{W} = \{c_1, c_2, ..., c_{n_w}\} \subset \mcal{C}.
\end{equation}
As the rover moves through the world and visits each waypoint, its energy level fluctuates.
Let the rover state ${\boldsymbol{x} \in \mcal{X}}$ be described by the tuple $<c,t,b,w>$, where:
\begin{itemize}
    \item $c \in \mcal{C} \subset \mbb{N} \times \mbb{N}$ is the terrain grid cell where the rover is located (e.g., a pixel on an orbital map);
    \item $t \in {\mbb{R}}_{\geq 0}$ is the time;
    \item $b \in {\mbb{R}}_{\geq 0}$ is the rover's battery energy, or state of charge (SOC);
    \item $w \in \mbb{N}$ is the index of the next waypoint in $\mcal{W}$ to visit.
\end{itemize}
We assume that $\mcal{C}$ only contains grid cells where it is safe for the rover to drive.
The state space is hybrid; state components $c$ and $w$ are discrete variables, and $t$ and $b$ are continuous variables.
Let the choice functions $c(\boldsymbol{x})$, $t(\boldsymbol{x})$, $b(\boldsymbol{x})$, and $w(\boldsymbol{x})$ represent the corresponding component of the state~$\boldsymbol{x}$.%

The action space $\mcal{A}$ includes mobility, waiting, and waypoint-specific actions.
From every state $\boldsymbol{x}$, up to eight mobility actions (driving to one of the adjacent grid cells) are available.
The duration of mobility actions depends on the terrain properties in the neighborhood of $c(\boldsymbol{x})$ and the change in SOC depends on the local illumination conditions when the action takes place.
Additionally, the rover can wait in place for a predetermined duration $\delta t_{\text{wait}}$.
In order to successfully ``visit'' a waypoint $c_k \in \mcal{W}$, a corresponding science action $a_k$ with a predetermined duration $\delta t_k$ and an energy cost $\delta b_k$ must be carried out.
This action is allowed only if ${c(\boldsymbol{x}) = c_k}$ and ${w(\boldsymbol{x}) = k}$.
In other words, the rover needs to be at the waypoint and it must have successfully visited all the previous ones in $\mcal{W}$.

As the rover performs its traverse, it is temporarily halted by random faults.
We assume that only one fault at a time can happen, and that faults occur at a known average spatial rate $\alpha$.
In practice, such knowledge can be inferred from previous mission data~\cite{robinson_intrepid_2020,keane_endurance_2022}.
We further assume that over disjoint drive segments, the fault probabilities are independent.
Thus, fault profiles can be modelled as a Poisson process~\cite[Chapter 2]{ross_stochastic_1995}.

Let $\rho(c,a)$ indicate the physical driving distance\footnote{This is the Euclidean distance in three-dimensional space between the centers of two neighbouring grid cells.} associated with taking action $a$ from grid cell $c$.
In theory, the number of faults occurring over this interval, represented by the random variable $F$, follows a Poisson distribution.
However, since the rover's autonomy software (or a ground operations team) can intervene after the occurrence of one fault, the forward dynamics of our model only account for whether an action terminates nominally (with probability that no fault occurs, or ${F=0}$), or gets interrupted by a fault (with probability that at least one fault occurs, or ${F \geq 1}$).
Respectively, these probabilities evaluate to:
\begin{equation}
\begin{aligned}
    \mathrm{Pr}(F = 0; c,a)    & =     \exp\big(\!\left.-\alpha\rho(c,a)\right.\!\big),\\
    \mathrm{Pr}(F \geq 1; c,a) & =  1 -\exp\big(\!\left.-\alpha\rho(c,a)\right.\!\big).    
\end{aligned}
\label{eq:fault_probs}
\end{equation}
Since fault occurrence is dependent on driving distance, only mobility actions are subject to faults.
All other (static) actions, for which the driving distance $\rho(c,a)=0$, always terminate nominally.

\begin{figure}[h]
    \centering
    \includegraphics[width=\columnwidth]{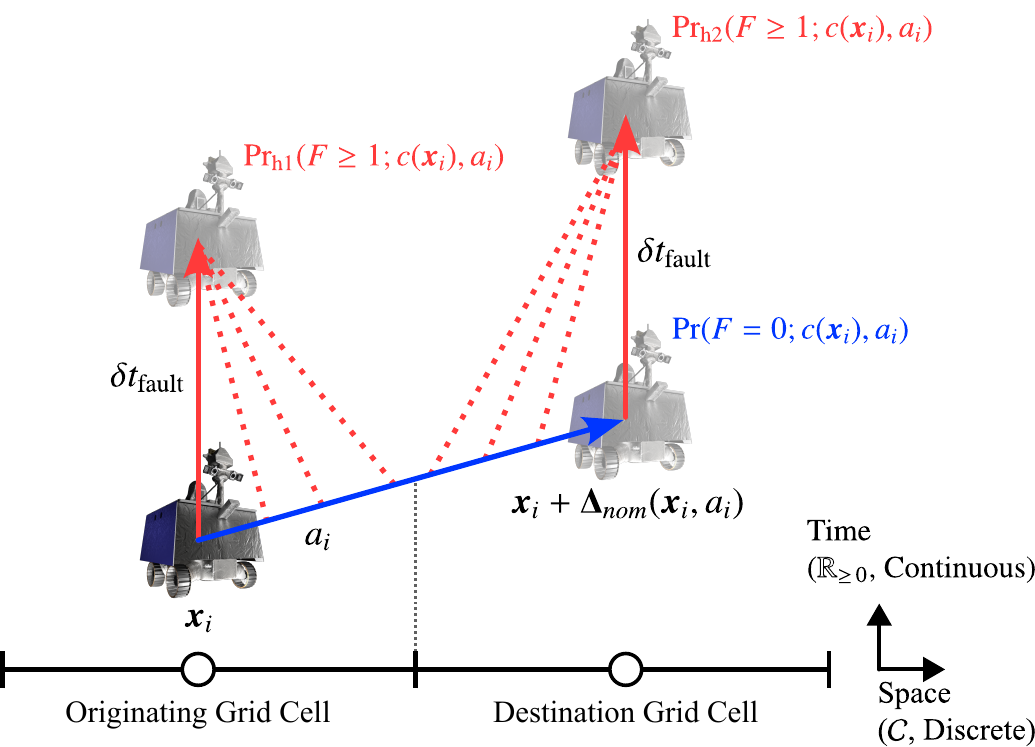}
    \caption{Conceptual view of the three possible state transitions and the associated probabilities for mobility action $a_i$ from originating state $\boldsymbol{x}_i$.
    For this state-action combination, the function $\boldsymbol{\Delta}_{nom}(\boldsymbol{x}_i, a_i)$ defines the change in state for a nominal transition.
    Only the spatial and temporal dimensions are visualized.%
    The nominal transition is shown in blue, while the fault-induced transitions are shown in red. 
    }
    \label{fig:transitions}
\end{figure}

When a fault interrupts a mobility action, there is a forced recovery period during which the rover stays in place for fixed time $\delta t_{\mathrm{fault}}$.
Although faults can occur anywhere along the drive interval, our rover model assumes navigation on a discrete grid; faults can either occur in the originating grid cell or in the destination grid cell.
\Cref{fig:transitions} shows the three possible outcomes associated with a mobility action. 
If a fault takes place in the first half of the drive interval, we approximate the state transition by a forced delay of $\delta t_{\mathrm{fault}}$ from the originating state.
This outcome has probability
\begin{equation}
    \text{Pr}_{\text{h1}}(F \geq 1; c,a) = 1 - \exp\big(-\alpha\rho(c,a)/2\big).
    \label{eq:half1_prob}
\end{equation}
Instead, if no faults occur in the first half of the drive interval but a fault does happen in the second half, the transition is approximated by a forced delay of $\delta t_{\mathrm{fault}}$ at the intended nominal terminal state, with probability
\begin{equation}
    \text{Pr}_{\text{h2}}(F \geq 1; c,a) = \exp\big(-\alpha\rho(c,a)/2\big)-\exp\big(-\alpha\rho(c,a)\big).
    \label{eq:half2_prob}
\end{equation}

\subsection{Safe Region of the State Space}

We adopt a definition of safety similar to the one provided in~\cite[Section 3.2]{lamarre_recovery_2023}, which solely depends on the rover's location, energy level and the time.
That is, the number of waypoints reached is irrelevant to rover safety.

Let $\mcal{H} \subset \mcal{C}$ be a predefined set of grid cells, which we call \textit{safe havens}, where the rover is allowed to terminate its traverse.
In the context of solar-powered mobility, safe havens are generally locations receiving, on average, substantial amounts of sunlight~\cite{shirley_overview_2022}.
For our work, we impose a safety criterion that also depends on the rover's energy and the time of arrival: upon arriving at a safe haven $h \in \mcal{H}$, the rover is `safe' if and only if it can hibernate in place to reach a minimum state of charge $\bar{b}_h$ by a predefined time $\bar{t}_h$.
The set of all safe states is denoted by $\mcal{S}$.
In practice, the design of this set (where the rover is allowed to terminate its traverse, with how much energy and by what time) would depend on mission needs after the traverse ends.

\subsection{Optimization Formulation}

Let $\mcal{O} \coloneqq \mcal{C} \times \mcal{T} \times \mcal{B}$ denote the (waypoint-agnostic) operational subset of the state space in which the rover must remain during its traverse.
The subset is bounded in time by the interval ${\mcal{T} = \left[t_{\text{min}},t_{\text{max}}\right]}$, representing the earliest and latest times during which the mission can occur.
A reasonable value for $t_{\text{max}}$ is the latest arrival time of all states in the safe set $\mcal{S}$.
The operational subset is also bounded in energy by the interval ${\mcal{B} = \left[b_{\text{min}},b_{\text{max}}\right]}$, representing the minimum acceptable SOC and the maximum one (for which a sensible choice would be the rover's battery capacity).
By definition, all safe states are in the operational set ($\mcal{S} \subset \mcal{O}$) and stepping outside of $\mcal{O}$ represents a mission failure.

We seek an online control law (a deterministic policy $\pi$) that maximizes the expected number of waypoints visited (a proxy for science return) from a given start state $\boldsymbol{x}_0 \in \mcal{O}$.
Let such a reward function be defined as
\begin{equation} \label{eq:reward}
    r(\boldsymbol{x}, a) =
    \begin{cases}
        1 & \text{if }c(\boldsymbol{x}) = c_k \in \mcal{W}, w(\boldsymbol{x}) = k, a=a_k, \\
        \zeta & \text{otherwise}, \\
    \end{cases}
\end{equation}
where $\zeta$ is a very small (negative) penalty.
Put into words, a positive reward is gained when taking a science action at the corresponding waypoint, while any other state-action combination results in a small penalty.
The policy must also respect an upper bound $\beta$ on the probability of mission failure, which occurs if the safe set is unreachable and exiting the operational region (i.e., running out of time or energy) is unavoidable.

\begin{problem}[CC Mission-Level Online Planning]
\label[problem]{prob:ccmlpp}
\begin{eqnarray}
    \max_{\pi \in \Pi} && \mbb{E}\left[\sum_{i=0}^{N-1} r(\boldsymbol{x}_i, \pi(\boldsymbol{x}_i))\right], \label{eq:obj} \\
    \mathrm{s.t.}  && \mathrm{Pr}\left\{ \bigwedge_{i=1}^{N} \boldsymbol{x}_i \in \mcal{X}_i \;\middle|\; \boldsymbol{x}_0,\pi \right\} \geq 1-\beta, \label{eq:cc} \\
    && \mcal{X}_i = \begin{cases}
        \mcal{S} & \text{if } i = N, \\
        \mcal{O} & \text{otherwise},\\
    \end{cases} \nonumber
\end{eqnarray}
\end{problem}
In Problem 3.1, $N$ is the number of steps taken until a terminal state is reached, which is always finite in this case.
In practice, we do not enforce a specific planning horizon and instead follow the policy until a termination condition is reached.\footnote{A terminal condition is reached in a finite number of steps since the operational time interval $\mcal{T}$ is bounded and all actions have a positive duration.
The maximum number of actions that can be taken in a trial is obtained by dividing the span of $\mcal{T}$ by the shortest action duration.}

\section{Risk-Bounded, Safe Mission-Level Path Planning}
\label{sec:method}

In essence, our approach leverages stochastic reachability to enforce risk bounds on the solutions obtained by existing mission-level path planning frameworks.
Drawing inspiration from existing AO-based, chance-constrained optimization methods, we construct \textit{partial} policy trees instead of complete ones for computational tractability reasons.
We first detail our policy tree generation technique and illustrate it with a relevant problem instance involving a single waypoint.
Then, we discuss incorporating this paradigm into TEMPEST, a risk-agnostic mission-level path planner introduced in \Cref{sec:relatedwork}, which is capable of planning through multiple waypoints over a state space similar to ours.
Lastly, as a solution to \Cref{prob:ccmlpp}, we detail a high-level online algorithm that relies on our risk-bounded implementation of TEMPEST.

\subsection{Risk-Bounded Optimization with Partial Policy Trees}

Existing risk-bounded AO optimization techniques conduct forward searches across finite sets of actions and outcomes up to some planning horizon~\cite{huang_hybrid_2018} or until a valid policy (an AO subtree whose leaf nodes all represent terminal states) is found~\cite{santana_rao_2016, gutow_andor_2022}.
In the worst-case, the number of nodes expanded increases exponentially with search depth.
Additionally, every expansion cycle is followed by a risk backpropagation step to all the ancestors of the newly-expanded nodes.
We propose the generation of \textit{partial} policy trees to alleviate the computational burden associated with standard risk-bounded AO search algorithms.
Conceptually, a partial policy tree is composed of two elements:
\begin{itemize}
    \item a nominal state trajectory (i.e., similar to a solution returned by a risk-agnostic offline planner) and all fault-induced state transitions branching off from it;
    \item a separate, pre-generated recovery policy dictating safe behavior from anywhere in the operational region of the state space, which we explain later.
\end{itemize} 
The creation of partial policy trees is enabled by two %
mechanisms: the (artificial) termination of OR nodes resulting from fault-induced state transitions, and backward search.
Respectively, these mechanisms effectively reduce the branching factor of the AO search, and only grow trees from which a target/safe region of the state space is within reach.
We explore these two mechanisms in detail below.

\subsubsection{OR Node Artificial Termination}

\begin{figure}[h]
    \centering
    \includegraphics[width=\linewidth]{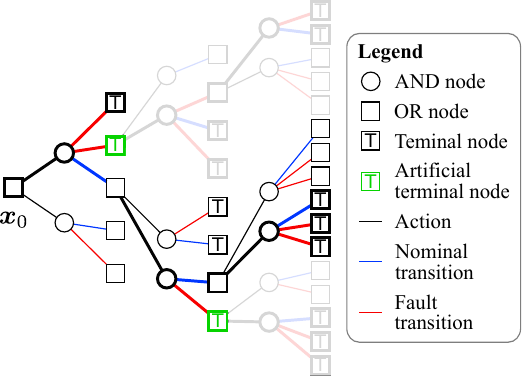}
    \caption{
        A partial policy tree (bold), grown from a given start state $\boldsymbol{x}_0$.
        OR nodes resulting from fault-related state transitions are artificially terminated (here, coloured in green).
        Without this mechanism, standard AO tree planners would return a complete policy, which would require many more expansion cycles (shown semi-transparent).
    }
    \label{fig:partial-tree}
\end{figure}

Partial policy trees are generated by solely expanding OR nodes associated with nominal state transitions.
Non-terminal OR nodes resulting from (usually low-probability) fault-induced state transitions are artificially terminated.
As conceptually illustrated in \Cref{fig:partial-tree}, a partial policy tree grown from a given start state $\boldsymbol{x}_0 \in \mcal{O}$ can be significantly smaller than the corresponding policy tree grown without artificial node termination. %

To estimate the risk at the root of any kind of policy tree, the probability of mission failure (or risk) at all terminal nodes is required.
For actual terminal nodes, this value is either 0 (if the corresponding state is in the safe region $\mcal{S}$) or 1 (if the corresponding state is outside of the operational region $\mcal{O}$).
For artificial terminal nodes, we rely on a pre-generated safe recovery policy $\pi_\mcal{S}$ like the one presented in~\cite{lamarre_recovery_2023}.
Such a policy maximizes the probability of reaching the safe region $\mcal{S}$ from a given state in the operational state space, ignoring all other objectives (such as the cost of growing the policy tree).
In~\cite{lamarre_recovery_2023}, given an adequate parametrization (such as a state space discretization with a sufficiently high resolution), the corresponding value function ${V_\mcal{S} : \mcal{O} \rightarrow \left[0,1\right]}$ conservatively predicts (i.e., does not underestimate) the risk from any state in the operational region.
The risk at the root node of the policy tree is then obtained by propagating all terminal node risk estimates to their ancestors.
In a partial policy tree, this risk value corresponds to the probability of mission failure if the rover follows the nominal traverse and then begins following the safe policy $\pi_\mcal{S}$ upon falling into a state that corresponds to an artificially-terminated node.

The computational speed improvements associated with partial policy tree search come at the expense of solution optimality.
Unlike optimal, complete policy trees, whose cost calculation requires enumerating all future state transition possibilities for a given control law, partial policy trees are solutions containing the fault-free traverse with the lowest cost.
However, unlike risk-agnostic planning, searching for partial policy trees allows the application of a chance-constraint, or an upper bound on the probability of mission failure.
We leave the study of the level of suboptimality associated with partial policy trees, which depends on factors like the policy depth and the probability of fault occurrence, as future work.

\subsubsection{Backward Search}
Because the execution risk from a node\footnote{Here, \textit{execution risk} is the probability of mission failure when following a policy tree from a given node.} depends on the risk of its (forward-time) successors, we propose performing a search from a target region of the state space backwards to a (given) start state.
When growing a partial policy tree, only nominal state transitions are back-propagated during a node expansion step. 
The execution risk from a newly-expanded node is calculated using the execution risk values of its immediate forward-time successors: the parent node from which it was expanded and the nodes corresponding to fault-induced state transitions.
A significant benefit of performing a backward search is that the execution risk from a (non-terminal) node does not change as the tree grows towards the start state.
When the search reaches the start state, it is trivial to verify whether a candidate partial policy tree respects a threshold on execution risk.
Additionally, as demonstrated in~\cite{huang_hybrid_2018}, policy trees respecting a risk threshold at their root node may still contain overly risky intermediate nodes along the corresponding nominal trajectory.
In this case, via backward search, dangerous nodes can be pruned before being added to the tree during the expansion phase.

\begin{figure}[!h]
    \centering
    \includegraphics[width=0.85\linewidth]{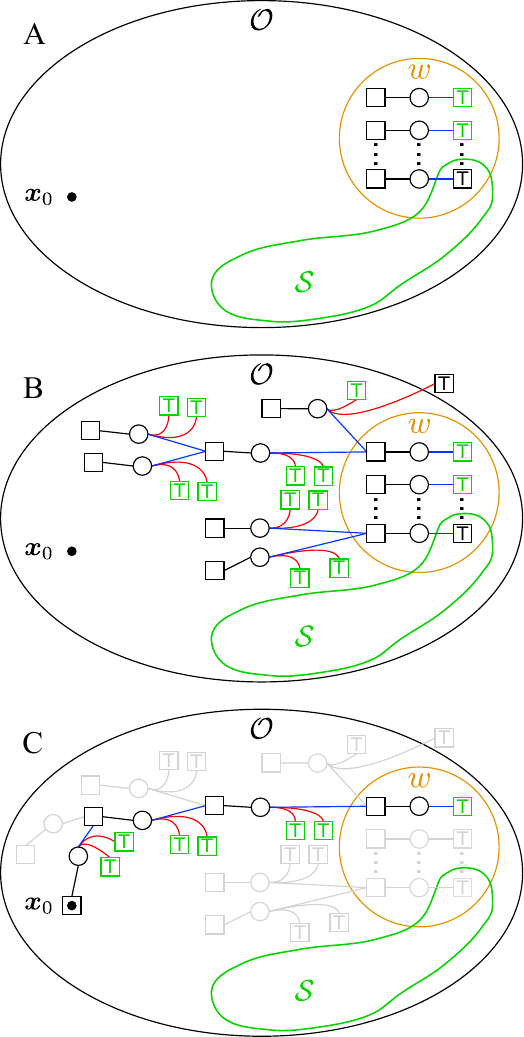}
    \caption{
        Backward search to find a risk-bounded partial policy tree.
        In forward-time, the tree guides the rover from a start state towards region $w$, where a waypoint action can be taken.
        Stage A consists of discretizing the waypoint region into terminal nodes and initializing the search.
        Stage B conducts the search according to a user-provided cost function, back-propagating nominal actions and calculating the risk at newly-created nodes using their immediate forward-time successors.
        Once the search reaches the start state (Stage C), the best partial policy tree is the forward-time branch leading to the waypoint region.
        Node and state transition colors follow the same convention as in \Cref{fig:partial-tree}.
    }
    \label{fig:backward-search}
\end{figure}

\Cref{fig:backward-search} depicts how a partial policy tree is be obtained through backward search in a problem instance conceptually similar to the one studied in this paper.
Let $\boldsymbol{x}_0$ be an initial rover state in the operational region $\mcal{O}$.
The orange set $w$ represents a region\footnote{This region represents states co-located at the grid cell of the corresponding waypoint, but with different time and energy values.} of the state space where a waypoint action can be taken once and provide a reward to the rover.
The green region $\mcal{S}$ represents the safe set of states, where we want the traverse to terminate with a certain probability.
The first step consists in discretizing the waypoint region into suitable terminal nodes (those that are outside of the safe set are artificially terminated).
Then, the search takes place according to a user-provided objective function.
Since we are growing partial policy trees, only nominal actions are propagated backwards in time during node expansion.
The risk at newly-expanded nodes is calculated using a one-step lookup of all possible outcomes in the forward-time direction; the risk from the parent node is already known, and the risk at all (artificially-terminated) nodes resulting from a fault-induced state transition are estimated using the safe policy $\pi_\mcal{S}$, which we introduced previously.
The best policy tree is the first branch that reaches the start state and satisfies the provided risk threshold.

\subsection{Risk-Bounded TEMPEST using Partial Policy Trees}

The single waypoint planning strategy previously described shares similarities with TEMPEST~\cite{tompkins_mission-level_2006}.
TEMPEST is a backward search algorithm that is capable of handling large state spaces and that can find state trajectories that visit multiple waypoints in a prespecified order.
We modify the traditional TEMPEST algorithm such that it generates (risk-bounded) partial policy trees spanning multiple waypoints instead of (risk-agnostic) paths.

With TEMPEST, multi-waypoint mission plans are found by breaking down the overall problem into individual search instances (called ``segments'') between consecutive waypoints.
As such, given an (ordered) list of $n_w$ waypoints to visit, $n_w$ search segments are created starting with the last one (due to the backward search methodology).
Since the reward function in~\eqref{eq:reward} is inherently maximized in this framework (TEMPEST attempts to find mission plans visiting every waypoint provided), the objective function followed by individual segments is application-dependent.
In our work, individual segments minimize the traverse time between consecutive waypoints.
If our risk-bounded TEMPEST algorithm fails to find a mission plan visiting all $n_w$ waypoints, it removes the last waypoint from the list and starts a new search for a plan that connects the remaining $n_w-1$ waypoints together.
This iterative shortening of the waypoint list continues until a feasible mission plan (i.e., a partial policy tree with a satisfactory execution risk) is found.
Otherwise, no plan to any waypoint exists.

\begin{figure}[h]
    \centering
    \includegraphics[width=\linewidth]{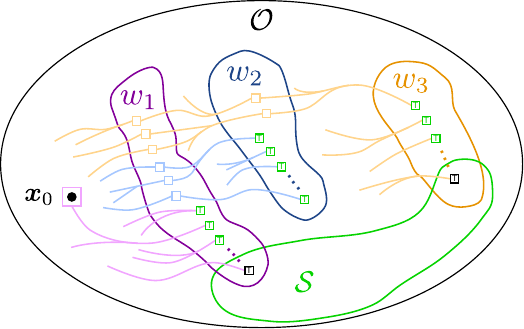}
    \caption{
        The iterative mechanism of risk-bounded TEMPEST.
        If no solution (feasible partial policy tree) is found for a given list of waypoints, the last waypoint is dropped and a new search is started with the updated waypoint list.
        Here, instances initialized in waypoint regions $w_3$ and $w_2$ fail to find feasible solutions in the search segment linking the start state to the first waypoint.
    }
    \label{fig:risk-tempest}
\end{figure}

\Cref{fig:risk-tempest} conceptually illustrates the iterative mechanism of risk-bounded TEMPEST searching for a feasible mission plan (a partial policy tree) visiting three waypoint regions in the operational subset $\mathcal{O}$.
First, a search is initialized in waypoint region $w_3$ and valid solutions are found for the two segments preceding it (the orange policy trees connecting $w_3$ to $w_2$, and those connecting $w_2$ to $w_1$) but no feasible solution is found beyond $w_1$.
Thus, the last waypoint ($w_3$) is dropped and a new search is initialized in waypoint region $w_2$.
Valid solutions are found for the segment preceding it (the blue policy trees between $w_2$ and $w_1$), but again no feasible solution is found beyond $w_1$.
Finally, the search initialized at the first waypoint ($w_1$) does find a feasible policy tree connecting to the start state.
We refer the reader to~\cite[Chapter 4.3]{tompkins_mission-directed_2005} for a thorough description of TEMPEST.
The details of our specific implementation (search mode and parametrization) are provided in \Cref{sec:experiments,sec:lcross}.

\subsection{High-Level Online Algorithm}

The risk-bounded TEMPEST planner discussed in the previous section finds partial policy trees from a given start state and through as many waypoints as possible.
The execution risk associated with these policies, which assume that the rover will start following the safe recovery policy $\pi_\mcal{S}$ after a fault occurs, provides an assurance that the chance-constraint imposed by~\eqref{eq:cc} is respected.
However, after a fault occurs, another feasible plan might exist from the current state (allowing the rover to accumulate a greater reward by visiting more waypoints).
As such, before abandoning the nominal traverse and following the recovery policy to the safe region, we simply re-query the risk-bounded TEMPEST algorithm from the current state.
Because of the backward-search mechanism of TEMPEST, only a search in the current segment is needed; existing search results from segments beyond the next waypoint are reused.
If no valid solution is found, then the rover simply follows the safe recovery policy until a terminal state is reached. 

\IncMargin{1em}
\begin{algorithm}[h]
\setstretch{1.25}
\SetAlgoNoLine\SetAlgoNoEnd
\caption{CC Mission-Level Online Path Planning}\label{alg:policy}
\KwInput{$\boldsymbol{x}_0$, $\pi_\mcal{S}$, $\mcal{W}$, $\beta$}

\Indp\Indpp %

$\boldsymbol{x} \gets \boldsymbol{x}_0$ \tcp{Initialize current rover state}

\BlankLine
\eIf{$V_\mcal{S}(\boldsymbol{x}) > \beta$}{
    \Indpp\Indpp %
    Return infeasible
}{
    \Indpp\Indpp %
    $\tau = \text{RiskBoundedTEMPEST}(\boldsymbol{x}, \mcal{W}, \beta)$ \label{line:policy:init}
}

\BlankLine
\tcp{Follow active partial policy tree}
\While{$\tau \neq$ null \textbf{and} $\boldsymbol{x} \neq \text{End}(\tau)$}{
    \Indpp\Indpp %
    $\boldsymbol{x} \gets \text{Step}(\boldsymbol{x}, \tau(\boldsymbol{x}))$

    \If{$\boldsymbol{x} \in \text{FaultNodes}(\tau)$}{
        \Indpp\Indpp %
        $\tau = \text{RiskBoundedTEMPEST}(\boldsymbol{x}, \mcal{W}, \beta)$
    }   
}

\BlankLine
\tcp{Follow safe recovery policy}
\While{\textbf{not} $\text{Terminal}(\boldsymbol{x})$}{
    \Indpp\Indpp %
    $\boldsymbol{x} \gets \text{Step}(\boldsymbol{x}, \pi_{\mcal{S}}(\boldsymbol{x}))$
}

\end{algorithm}

We illustrate this high-level online planning strategy in \Cref{alg:policy}.
Lines 2-5 verify whether a feasible solution exists and, in the positive case, an initial partial policy tree $\tau$ is generated for the start state.
In lines 6-9, the rover follows the currently-active policy tree as long as it exists and it has not been fully traversed.
Every cycle involves updating the current rover state by stepping through the world according to the best policy tree action $\tau(\boldsymbol{x})$.
If a fault occurs and the current rover state corresponds to one of the artificially-terminated nodes resulting from a fault-induced state transition), the active policy tree is updated.
Lastly, in lines 10-11, the rover follows the safe recovery policy until a terminal condition is reached (either the safe set $\mcal{S}$ is entered or it becomes unreachable).

\section{Simulation Experiments}
\label{sec:experiments}

We demonstrate our approach with simulated kilometre-scale traverses using orbital maps of Cabeus Crater at the lunar south pole.
Our aim is to empirically characterize the behaviour of \Cref{alg:policy} on a medium-scale problem before focusing on a long-range driving scenario near the LCROSS impact site (discussed in the next section).

For a given mission profile, we begin by highlighting the dangers associated with risk-agnostic mission-level path planning and illustrating the safety benefits of our risk-bounded approach.
Then, we carry out Monte Carlo trials using random fault profiles to demonstrate the ability of our method to maximize the number of waypoints visited while respecting a chance constraint.
The results are compared against a baseline provided by an existing chance-constrained planning algorithm that uses dynamic programming.\footnote{Supplementary results, plots and animations are available at \href{https://papers.starslab.ca/safe-mission-level-planning}{https://papers.starslab.ca/safe-mission-level-planning}}%

\subsection{Environment and Rover Models}

\begin{figure}[h]
    \centering
    \includegraphics[width=\columnwidth]{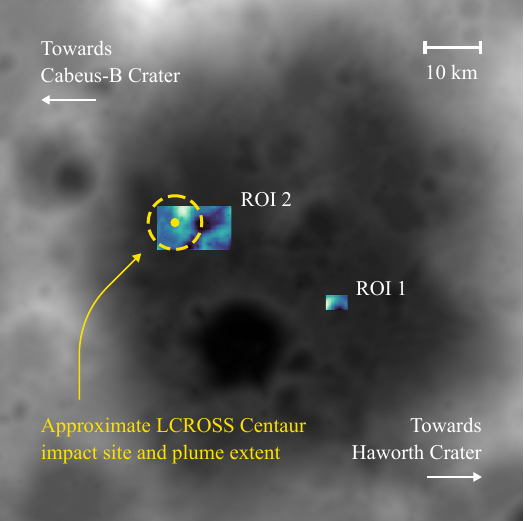}
    \caption{Elevation model of Cabeus Crater (lighter is higher). Directions to nearby craters are indicated to provide a sense of orientation. The submaps employed in \Cref{sec:experiments,sec:lcross} (ROI 1 and 2, respectively) are highlighted. The LCROSS impact site location is inferred from figures in~\cite{schultz_lcross_2010}.}
    \label{fig:orbital_map}
\end{figure}

We use a dataset of solar illumination maps of the Cabeus Crater area created by the NASA Jet Propulsion Laboratory.
Each georeferenced map indicates the fraction of the solar disk visible two metres above the lunar surface on an hourly basis.
Data are expressed in 20\% increments and cover a period of three lunar synodic days (August 1 to October 27, 2029).
The maps have a spatial resolution of 240 metres per pixel and account for possible terrain occlusions within the spatial extent of the dataset and beyond.
Georeferenced terrain elevation information, from which slope and aspect (slope azimuth) maps are calculated, is retrieved from a high-resolution mosaic of the lunar south pole composed of Chang'e 2 images~\cite{li_lunar_2018}.
The terrain maps are reprojected to a resolution of 240 metres per pixel to match the resolution of the solar illumination dataset.
\Cref{fig:orbital_map} shows the regions of interest for the experiments in this paper within Cabeus Crater.
Low-level orbital data queries, such as the calculation of driving distances and solar energy generation over time intervals, are handled by the gplanetary\_nav Python package.\footnote{The package is available at \href{https://github.com/utiasstars/gplanetary-nav}{https://github.com/utiasstars/gplanetary-nav}}

Rover model parameters are shown in \Cref{tab:exp1_rover_model}.
Mobility is restricted to an eight-connected grid pattern according to the pixel grid on the orbital maps.
Cells/pixels corresponding to a terrain slope magnitude of more than 20 degrees are avoided.
The duration and energy cost of a drive action between adjacent grid cells depend on the local terrain conditions. %
Although slip and surface rock density (CFA) models can easily be incorporated when calculating these values, here we employ a constant drive velocity and power model for simplicity; driving time and energy costs are solely dependent on the physical distance between adjacent grid cells.

To estimate solar power, we assume that the rover maintains a constant area of solar panels perfectly oriented towards the Sun at all times, mimicking panels mounted on a pan-tilt platform.
When the full solar disk is visible (i.e., when there are no terrain occlusions), we use a constant solar irradiance of 1,367~W/m\textsuperscript{2}.
We keep the incorporation of more complex power generation models as future work.

\begin{table}[t]
    \centering
    \caption{Medium-scale mission rover model.}
    \begin{tblr}{
    cells={l},
    row{1} = {rowsep=3pt}
}
\toprule
\textbf{Parameter} & 
\textbf{Value}\\
\midrule
\TCategoryAboveSpace
Solar panel area                        & 1.5 m\textsuperscript{2} \\
Solar panel efficiency                  & 30 \% \\
Driving velocity                        & 0.05 m/s \\
Driving power draw                      & 110 W \\
Fault resolving power draw              & 80 W \\
Idling power draw (waiting in place)    & 80 W \\
Idling power draw (hibernating)         & 30 W \\
Battery capacity                        & 7,000 Wh \\
[2mm] \bottomrule       
\end{tblr}
    \label{tab:exp1_rover_model}
\end{table}

\subsection{Medium-Scale Mission Scenario}

\begin{figure}[h]
    \centering
    \includegraphics[width=\columnwidth]{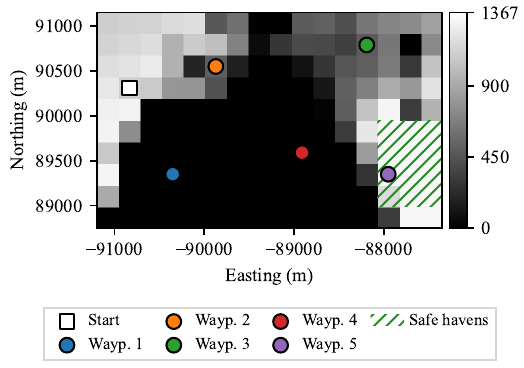}
    \caption{
        Average solar irradiance map (W/m\textsuperscript{2}) during the operational time interval over ROI 1 shown in \Cref{fig:orbital_map}.
        Every pixel is 240 metres wide.
        The rover's start location and the positions of waypoints are indicated.
        The hatch pattern marks the locations of safe havens (i.e., where the traverse is allowed to terminate).
        }
    \label{fig:exp1_setup}
\end{figure}

Initially, we focus on a medium-scale traverse scenario using the submap labelled ``ROI 1'' in \Cref{fig:orbital_map}.
A close-up view of this region is provided in \Cref{fig:exp1_setup}, showing the average solar irradiance during the operational time interval of the proposed mission.
Key locations, such as the rover's starting position and the locations of waypoints to visit, are marked.
Waypoints 1 and 4 are in the shade over the whole duration of this experiment.
The parameters and requirements of the proposed mission are detailed in \Cref{tab:exp1_mission_params}.
In addition to mobility and waypoint-specific actions, the rover can wait in place for a duration of $\delta t_{\text{wait}}$ = 1,800 seconds (30 minutes).

The hatch pattern in \Cref{fig:exp1_setup} indicates the safe region where the rover must terminate its traverse.
In this case, upon arrival at any of the safe havens, the rover must be able to hibernate in place such that its energy level is at least 2,000~Wh by the end of the operational time interval.
Unlike our approach, the risk-agnostic planner we compare our algorithm against requires an end location inside of the safe region.
Therefore, we place the last waypoint (i.e., 5) at one of the safe havens.

\begin{table}[h]
    \centering
    \caption{Medium-scale mission parameters and requirements.}
    \begin{tblr}{
    row{1} = {rowsep=3pt}
}
\toprule
\textbf{Parameter}          & \SetCell[c=2]{l}\textbf{Value} \\
\midrule
\TCategoryAboveSpace
\textbf{Operational subset $\mathcal{O}$} & & \\
Start time                  & \SetCell[c=2]{l} Aug 30 2029 12:33:20 \\
End time                    & \SetCell[c=2]{l} Sep 2 2029 22:33:20 \\
Energy interval             & \SetCell[c=2]{l} 500 Wh to 7,000 Wh \\
\TCategoryAboveSpace
\textbf{Safe subset $\mathcal{S}$} & & \\
Safe haven locations        & \SetCell[c=2]{l} (See \Cref{fig:exp1_setup}) \\
Time limit                  & \SetCell[c=2]{l} Operational end time \\
Min. energy                 & \SetCell[c=2]{l} 2,000 Whr at time limit \\
\TCategoryAboveSpace
\textbf{Random fault model} & & \\
Average fault rate          & \SetCell[c=2]{l} 1 every 5,000 m driven \\
Fault recovery duration     & \SetCell[c=2]{l} 36,000 s (10 hours) \\
\TCategoryAboveSpace
\textbf{Initial State}      & & \\
Time                        & \SetCell[c=2]{l} Aug 30 2029 12:33:20 \\
Battery energy              & \SetCell[c=2]{l} 1,000 Wh \\
\TCategoryAboveSpace
\textbf{Waypoint actions}   & \textbf{Duration} & \textbf{Energy cost} \\
Waypoint 1                  & 7,200 s           & 2,000 Wh \\
Waypoint 2                  & 7,200 s           & 2,000 Wh \\
Waypoint 3                  & 10,800 s          & 3,000 Wh \\
Waypoint 4                  & 7,200 s           & 1,500 Wh \\
Waypoint 5                  & 7,200 s           & 1,000 Wh \\
[2mm] \bottomrule
\end{tblr}
    \label{tab:exp1_mission_params}
\end{table}

\subsection{Initial Mission Plans}

We highlight the safety benefits of chance-constrained optimization by generating two initial traverse plans:
\begin{enumerate}
    \item a nominal mission-level plan employing standard (risk-agnostic) TEMPEST;
    \item an initial partial policy tree using our risk-bounded implementation, which corresponds to \Cref{line:policy:init} in \Cref{alg:policy}.
\end{enumerate}
Both techniques use the same baseline TEMPEST configuration, which is detailed in \Cref{tab:exp1_algorithm_params}.
Parameters of the safe recovery policy, which is used to calculate risk, are also included.
We refer the reader to~\cite[Chapter 4.3]{tompkins_mission-directed_2005} for an in-depth description of the TEMPEST algorithm %
and to~\cite{lamarre_recovery_2023} for details regarding the safe recovery policy.

\begin{table}[h]
    \centering
    \caption{Risk-bounded TEMPEST algorithm parameters for the medium-scale mission scenario.}
    \begin{tblr}{
    row{1} = {rowsep=3pt}
}
\toprule
\textbf{Parameter}                  & \textbf{Value}\\
\midrule
\TCategoryAboveSpace
\textbf{Baseline TEMPEST}           & \\
Time class resolution               & 1,800 s \\
Energy class resolution             & 150 Wh \\
Objective function                  & Traverse time \\
Max. iterations per search          & 15,000 \\
Resolution-equivalence pruning      & Lower energy is better \\
State dominance pruning             & None \\
\TCategoryAboveSpace
\textbf{Risk-bounded TEMPEST}       & \\
Risk bound ($\beta$)                & 2\% everywhere \\
\TCategoryAboveSpace
\textbf{Safe recovery policy}       & \\
Time discretization resolution      & 1,800 s\\
Energy discretization resolution    & 150 Wh\\
Convergence criterion               & 1e-5 \\
[2mm] \bottomrule
\end{tblr}
    \label{tab:exp1_algorithm_params}
\end{table}

Our implementation of TEMPEST is akin to the original algorithm using the Incremental Search Engine (ISE) in the ``BESTPCOST'' mode.
The search takes place across a four-dimensional hybrid state space (the mobility grid and the continuous time and energy dimensions) and minimizes the traverse time between consecutive waypoints.
However, since we are not interested in dynamic plan repair, our implementation instead relies on a heuristic A* search mechanism, which provides the same result as the initial iteration of ISE.
When two nearby states in the search tree fall into the same equivalence class (called ``resolution equivalence'' in the original work), the one with the lowest energy requirement is kept.

\subsubsection{Risk-Agnostic Mission Plan}

\begin{figure}[b!]
    \centering
    \includegraphics[width=\columnwidth]{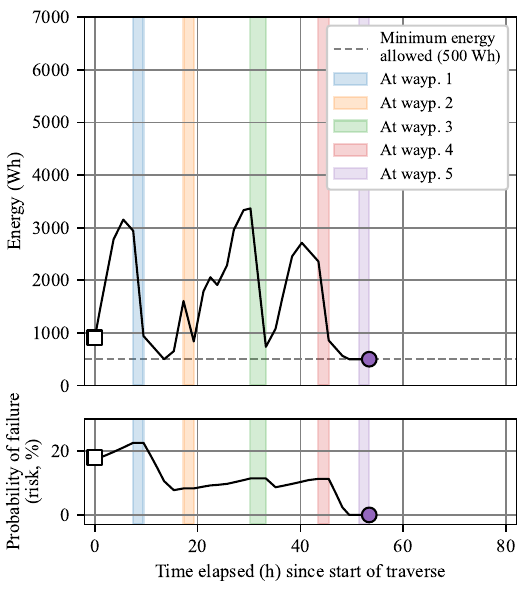}
    \caption{\textit{Top}: Minimum energy required as a function of time along the nominal mission plan found with standard (risk-agnostic) TEMPEST.
    The total duration of the plan is 53.4 hours, which is within the 82-hour operational time interval.
    \textit{Bottom}: Estimate of the joint risk along the mission plan as a function of time.
    The most dangerous portion of the traverse is near the first waypoint, where risk reaches a maximum of 23\%.
        }
    \label{fig:exp1_std_energyrisk}
\end{figure}

\begin{figure}[!h]
    \centering
    \includegraphics[width=\columnwidth]{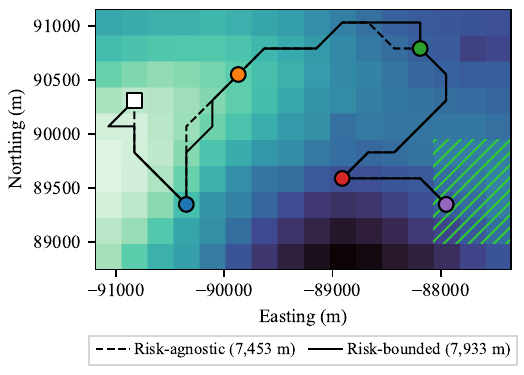}
    \caption{
        Paths found by standard (risk-agnostic) and risk-bounded TEMPEST overlaid onto the elevation map of ROI 1.
        The risk-bounded traverse is roughly half a kilometre longer.
        Significance of markers and hatch pattern is the same as in \Cref{fig:exp1_setup}.
        }
    \label{fig:exp1_paths}
\end{figure}

The result returned by standard TEMPEST, expressed as a sequence of spatiotemporal states and a minimum energy requirement for each state, is shown in \Cref{fig:exp1_paths} and at the top of \Cref{fig:exp1_std_energyrisk}.
The former illustrates the physical path proposed and the latter shows the corresponding minimum energy curve as a function of the time elapsed from the start of the traverse.
Although this solution was generated using a risk-agnostic approach, we can still estimate the joint probability of failure along this mission plan using the safe recovery policy.
Shown at the bottom of \Cref{fig:exp1_std_energyrisk}, the risk curve peaks at 23\% when the rover is near the first waypoint.
One of the reasons the rover is so vulnerable in this section of the traverse is the (arguably unacceptably) low allowed energy level, despite being far from any source of sunlight and from the safe region.
The risk in other sections of the mission plan, which threads around 10\%, is another contributing factor.

\subsubsection{Risk-Bounded Mission Plan}
\begin{figure}[t]
    \centering
    \includegraphics[width=\columnwidth]{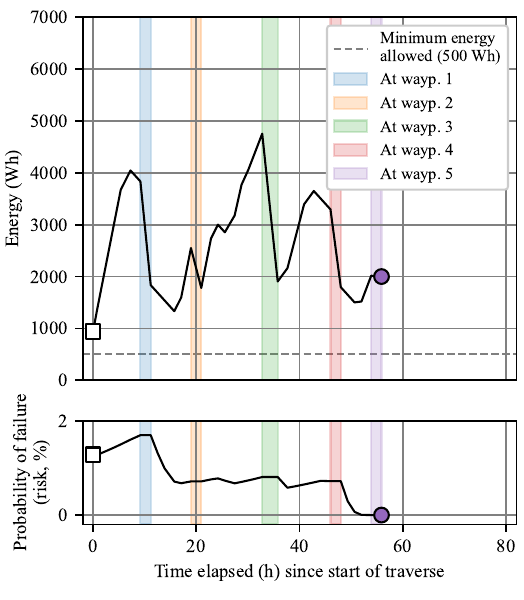}
    \caption{\textit{Top}: Minimum energy required as a function of time along the nominal traverse of the partial policy tree found by risk-bounded TEMPEST.
    The total duration of the plan is 55.9 hours, which is within the 82-hour operational time interval.
    \textit{Bottom}: Joint risk associated with the traverse, which remains within the imposed 2\% bound at all times.
    The most dangerous period of the traverse is at the first waypoint, where risk reaches a maximum of 1.7\%.
    }
    \label{fig:exp1_rb_energyrisk}
\end{figure}

We impose a bound on the probability of failure $\beta$ = 2\% everywhere along the traverse, which is an order of magnitude lower than the most dangerous section of the risk-agnostic solution.
The nominal traverse of the resulting partial policy tree, expressed in the same way as the risk-agnostic solution, is plotted in \Cref{fig:exp1_paths,fig:exp1_rb_energyrisk}.
Compared with the risk-agnostic solution, the risk-bounded traverse is half a kilometre longer and takes an additional two hours, roughly, to complete.
The traverse schedule differs slightly from the risk-agnostic one, maintaining the rover's exposure to sunlight slightly longer and keeping it within proximity of safety where and when it matters.
The entire joint risk profile stays within the imposed 2\% bound.

\subsection{Monte Carlo Simulations}

We now show the ability of \Cref{alg:policy} to recover from faults and to maximize the total science reward (i.e., the number of waypoints visited) while respecting an upper bound on risk (as before, $\beta$ = 2\%).
We simulate 10,000 Monte Carlo traverse experiments, each affected by a different random fault profile sampled from the underlying Poisson process (for the fault model parameters listed in \Cref{tab:exp1_mission_params}).
To successfully visit a waypoint (and receive a reward of +1), the corresponding waypoint action must be successfully completed.
Any other state-action combination results in the small penalty $\zeta$ =-0.001.
A trial that ends at a state in the safe region $\mathcal{S}$ is considered to be a success.
Any trial in which the rover exits the operational region of the state space $\mathcal{O}$ or moves out of reach of the safe region (i.e., when the safe recovery policy predicts a risk of 100\%) is considered to be a failure.
The actual risk associated with our algorithm is, approximately, the ratio of failed trials.

We compare the performance of our approach against a policy generated with the chance-constrained dynamic programming (CCDP) algorithm presented in~\cite{ono_chance-constrained_2015}.
Although the CCDP algorithm converges to the optimal solution for a given state space discretization, it is usually much more computationally expensive than our approach due to its reliance on dynamic programming.
To clarify, our algorithm also partially relies on dynamic programming to generate the safe recovery policy (only) that is used to estimate risk.
However, since the CCDP algorithm acts as the main chance-constrained mission-level online planning algorithm, it must operate over a state space that includes a dimension to keep track of the number of waypoints visited.
Since we assume the waypoints must be visited in a specific order, the added state space size, compared to that of the safe recovery policy, is linear in the total number of waypoints.
In the current experiment, the size of the state space associated with the safe recovery policy is 1,154,561 while the one associated with the CCDP algorithm is 6,927,361.
In addition to managing the larger state space, the CCDP algorithm is iterative in nature---dynamic programming has to be carried out multiple times until convergence.

Furthermore, the CCDP algorithm tends to be conservative.
One of the contributing factors is that it uses a conservative map between the hybrid space (in which the simulation takes place) and the discretized representation (which the CCDP algorithm uses).
As documented in~\cite{lamarre_recovery_2023}, a conservative mapping mitigates dangerous risk prediction errors from a given start state.
To provide a fair comparison, the same mapping function and discretization resolutions are employed for the safe recovery policy.

We carried out our experiment on a server running Ubuntu 18.04 with an AMD Threadripper 2920X 3.5 GHz 12-Core CPU and 128 GB of memory.
Despite parallelization, the CCDP policy took more than two hours to produce.
Generating the initial policy tree for our algorithm (including the safe recovery policy), on the other hand, took a fraction of the time (30 minutes).
Each Monte Carlo trial took from a few seconds to a few minutes to roll out (using our approach).

A summary of the results is shown in \Cref{tab:exp1_MC_results}.
Both algorithms respect the imposed risk bound of 2\%, though the CCDP policy is more conservative than our algorithm.
Nevertheless, the mean reward achieved with our approach is very close to that of the CCDP algorithm, empirically demonstrating that we are able to maximize the number of waypoints visited.

\begin{table}[h]
    \centering
    \caption{Results from 10,000 Monte Carlo trials.}
    \begin{tblr}{
    colspec={p{0.17\columnwidth}p{0.13\columnwidth}p{0.15\columnwidth}p{0.27\columnwidth}},
    row{1} = {rowsep=3pt}
}
\toprule
\textbf{Algorithm}  & \textbf{Mean reward}  & \textbf{Actual risk (\%)} & \textbf{Time to initial solution (s)} \\
\midrule
\TCategoryAboveSpace
CCDP    & 2.79  & 0.8   & 7,315 ($\approx$ 2 h)  \\
Ours    & 2.77  & 1.5   & 1,889 ($\approx$ 0.5 h) \\
[2mm] \bottomrule
\end{tblr}
    \label{tab:exp1_MC_results}
\end{table}

\section{Risk-Bounded Traverses Near LCROSS}
\label{sec:lcross}

\begin{figure}[h]
    \centering
    \includegraphics[width=\columnwidth]{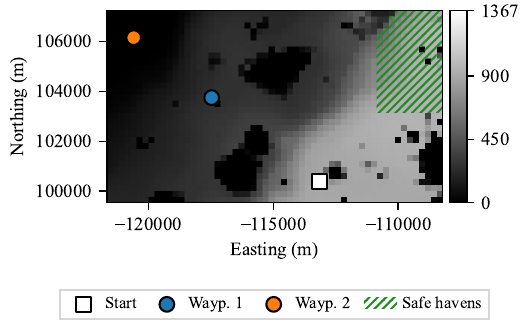}
    \caption{
        Average solar irradiance map (W/m\textsuperscript{2}) during the operational time interval over ROI 2 shown in \Cref{fig:orbital_map}.
        This region measures 7.7 kilometres by 13.4 kilometres wide.
        The rover's start location and the position of waypoints are indicated.
        The hatch pattern marks the location of safe havens (i.e., where the traverse is allowed to terminate).
        }
    \label{fig:exp2_setup}
\end{figure}

We simulate a long-range traverse using data from ``ROI 2'' in \Cref{fig:orbital_map}, which is in the vicinity of the LCROSS impact site.
The average solar irradiance over this region for the duration of the proposed mission is shown in \Cref{fig:exp2_setup}.
The waypoints are located on the left side of the map, inside of the LCROSS impact region, and the rover start location and end region (safe havens) are in places that are relatively more illuminated, on average.
The first waypoint is in a region receiving sunlight for a brief period of time and the second waypoint is constantly in the shade and possibly inside of a PSR.

Similar to the previous experiment, the rover %
stays away from terrain with a slope magnitude that exceeds 20 degrees.
It is also allowed to take wait actions for $\delta t_{\text{wait}}$ = 3,600 seconds (1 hour).
All rover model parameters for this experiment are listed in \Cref{tab:exp2_rover_model}.

\begin{table}[h]
    \centering
    \caption{Large-scale mission rover model.}
    \begin{tblr}{
    cells={l},
    row{1} = {rowsep=3pt}
}
\toprule
\textbf{Parameter} & 
\textbf{Value}\\
\midrule
\TCategoryAboveSpace
Solar panel area                        & 1.5 m\textsuperscript{2} \\
Solar panel efficiency                  & 30 \% \\
Driving velocity                        & 0.1 m/s \\
Driving power draw                      & 300 W \\
Fault resolving power draw              & 50 W \\
Idling power draw (waiting in place)    & 40 W \\
Idling power draw (hibernating)         & 30 W \\
Battery capacity                        & 30,000 Wh \\
[2mm] \bottomrule       
\end{tblr}
    \label{tab:exp2_rover_model}
\end{table}

The requirements for this large-scale mission scenario are listed in \Cref{tab:exp2_mission_params}.
The traverse time limit is the same for the entire safe region (August 31 2029, at 20:33:20).
The energy required at this time limit, however, varies from one safe haven to the next; the rover must be able to hibernate in place until the \textit{next} lunar day and reach a SOC of 50\% (15,000 Wh) by September 26 2029, at 17:33:20.
This implies that to successfully (and safely) terminate the traverse, the rover must have enough energy to survive the lunar night that follows. %
Additionally, we require that the first waypoint be visited while it is illuminated, which happens over a 34 hours-long time window (from August 26 2029 at 08:33:20 until August 27 2029 at 18:33:20).
Such a requirement simulates activities like panorama acquisition or any other scientific observation tasks that requires sunlight.

\begin{figure*}[t]
    \centering
    \includegraphics[width=\textwidth]{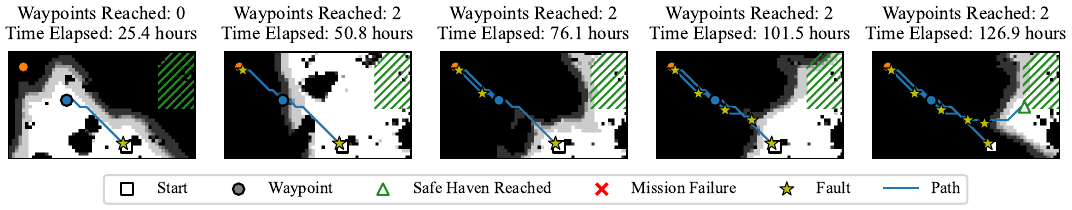}
    \caption{
        Traverse simulation where the rover visits all waypoints and reaches safety despite being delayed by six faults.
        For each frame, the background map illustrates the instantaneous illumination conditions.
        }
    \label{fig:exp2_traverse0}
\end{figure*}

\begin{figure*}[t]
    \centering
    \includegraphics[width=\textwidth]{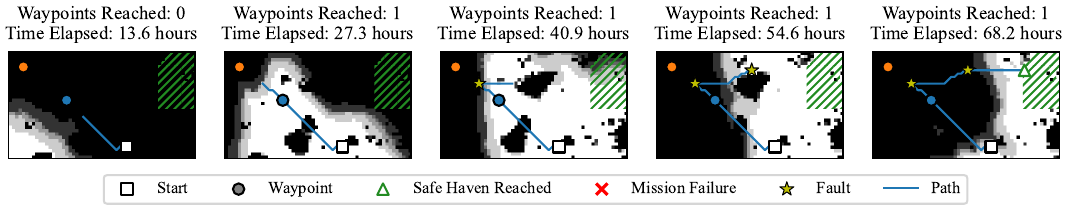}
    \caption{
        Traverse simulation where the rover visits only the first waypoint and successfully exits to safety shortly afterwards.
        Two faults were experienced.
        For each frame, the background map illustrates the instantaneous illumination conditions.
        }
    \label{fig:exp2_traverse1}
\end{figure*}

\begin{figure*}[t]
    \centering
    \includegraphics[width=\textwidth]{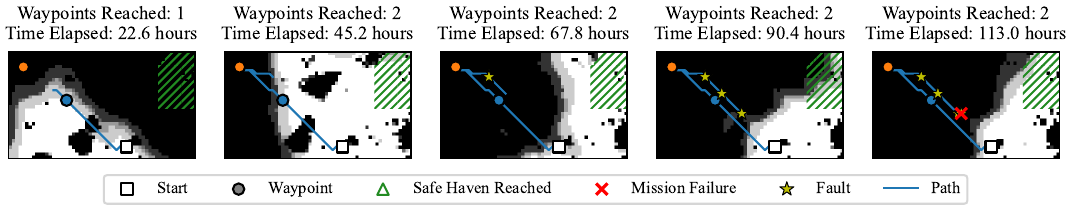}
    \caption{
        Traverse simulation where the rover visits all waypoints, but falls out of reach of the safe region after being delayed by six faults.
        For each frame, the background map illustrates the instantaneous illumination conditions.
        }
    \label{fig:exp2_traverse2}
\end{figure*}

We generate risk-bounded online traverse strategies subject to a maximum probability of failure of 5\%.
Relevant simulation and algorithmic parameters are shown in \Cref{tab:exp2_algorithm_params,tab:exp2_mission_params}.
\Cref{fig:exp2_traverse0,fig:exp2_traverse1} shows trials that visit the second and first waypoints, respectively, and then that reach safety.
\Cref{fig:exp2_traverse2} illustrates an instance where the rover fails to reach the safe region after successfully visiting both waypoints.

\begin{table}[h]
    \centering
    \caption{Large-scale mission parameters and requirements.}
    \begin{tblr}{
    row{1} = {rowsep=3pt}
}
\toprule
\textbf{Parameter}          & \SetCell[c=2]{l}\textbf{Value} \\
\midrule
\TCategoryAboveSpace
\textbf{Operational subset $\mathcal{O}$} & & \\
Start time                  & \SetCell[c=2]{l} Aug 25 2029 14:33:20 \\
End time                    & \SetCell[c=2]{l} Aug 31 2029 20:33:20 \\
Energy interval             & \SetCell[c=2]{l} 500 Wh to 30,000 Wh \\
\TCategoryAboveSpace
\textbf{Safe subset $\mathcal{S}$} & & \\
Safe haven locations        & \SetCell[c=2]{l} See \Cref{fig:exp2_setup} \\
Time limit                  & \SetCell[c=2]{l} Operational end time \\
Min. energy                 & \SetCell[c=2]{l} Variable (refer to text) \\
\TCategoryAboveSpace
\textbf{Random fault model} & & \\
Average fault rate          & \SetCell[c=2]{l} 1 every 5,000 m driven \\
Fault recovery duration     & \SetCell[c=2]{l} 36,000 s (10 hours) \\
\TCategoryAboveSpace
\textbf{Initial State}      & & \\
Time                        & \SetCell[c=2]{l} Aug 25 2029 14:33:20 \\
Battery energy              & \SetCell[c=2]{l} 25,000 Wh \\
\TCategoryAboveSpace
\textbf{Waypoint actions}   & \textbf{Duration} & \textbf{Energy cost} \\
Waypoint 1                  & 3,600 s           & 1,000 Wh \\
Waypoint 2                  & 14,400 s          & 3,000 Wh \\
[2mm] \bottomrule
\end{tblr}
    \label{tab:exp2_mission_params}
\end{table}

\begin{table}[h!]
    \centering
    \caption{Risk-bounded TEMPEST algorithm parameters for the large-scale mission scenario.}
    \begin{tblr}{
    row{1} = {rowsep=3pt}
}
\toprule
\textbf{Parameter}                  & \textbf{Value}\\
\midrule
\TCategoryAboveSpace
\textbf{Risk-bounded TEMPEST}           & \\
Time class resolution               & 3,600 s \\
Energy class resolution             & 250 Wh \\
Objective function                  & Traverse time \\
Max. iterations per search          & 100,000 \\
Resolution-equivalence pruning      & Lower energy is better \\
State dominance pruning             & None \\
Risk bound ($\beta$)                & 5\% everywhere \\
\TCategoryAboveSpace
\textbf{Safe recovery policy}       & \\
Time discretization resolution      & 1,800 s\\
Energy discretization resolution    & 200 Wh\\
Convergence criterion               & 1e-5 \\
[2mm] \bottomrule
\end{tblr}
    \label{tab:exp2_algorithm_params}
\end{table}


\section{Conclusion}
\label{sec:conclusion}

We have described a chance-constrained optimization problem whose solution yields safe exploration plans for the lunar south pole by a solar-powered rover.
Our proposed approach combines conventional mission-level path planning with stochastic reachability to generate online traverse strategies.
The policy guides the rover to waypoints of scientific interest while maintaining mission risk levels below a pre-specified threshold.
Through simulated kilometre-scale traverses using orbital terrain and illumination maps of Cabeus Crater, we empirically validated the performance of our algorithm.
Monte Carlo experiments showed that our approach is able to maximize the number of waypoints visited while ending traverses in a safe state. %
This work is relevant to both autonomous strategic and global mobility planning and human operator assistance.

The present work opens the door to numerous other research opportunities.
For instance, our approach remains sensitive to the size of the search space, since it requires the use of dynamic programming to calculate a safe recovery policy and heuristics to maintain reasonable performance with the mission-level path planner.
Improved memory and computational performance would make it possible to generate very long-range mission plans similar to those presented in~\cite{robinson_intrepid_2020,keane_endurance_2022}, or medium-scale plans using high-resolution orbital maps.
Also, relaxing the assumption of a single fault rate and a constant recovery time would be beneficial: in planetary exploration scenarios, issues multiple rover subsystems can halt a drive and faults have varying degrees of severity.

\section*{Acknowledgments}

The work of Shantanu Malhotra was supported by Jet Propulsion Laboratory, California Institute of Technology, under a contract with the National Aeronautics and Space Administration (80NM0018D0004).
We also thank the NASA Public Affairs Office for providing us with a high-resolution rendering of the VIPER rover and the NASA Jet Propulsion Laboratory, California Institute of Technology, for letting us use their lunar south pole solar illumination dataset.

\bibliographystyle{ieee/IEEEtran}
\bibliography{bibtex/robotics_abbrv.bib, bibtex/refs_strip.bib}

\begin{biographywithpic}
{Olivier Lamarre}{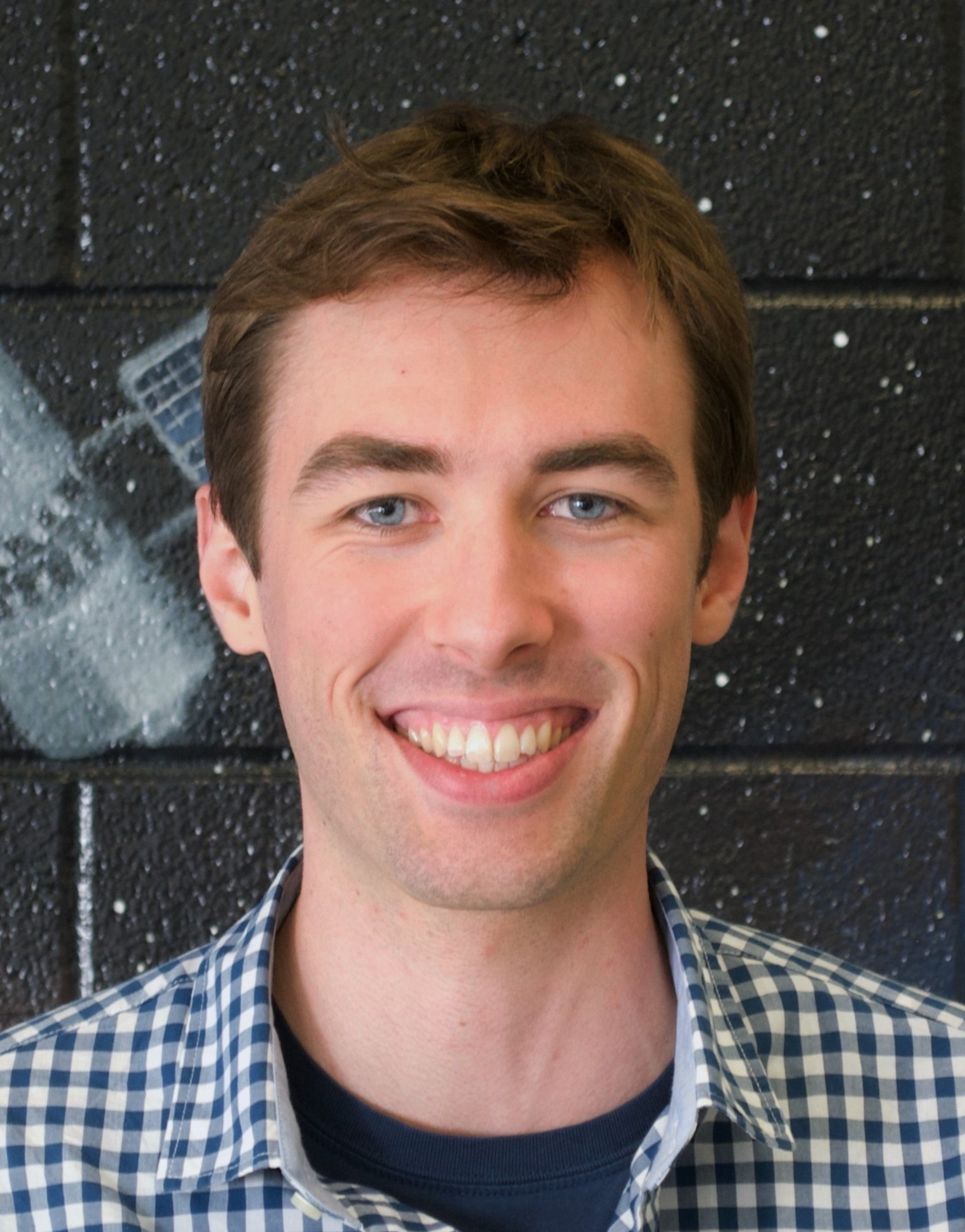}
is a Ph.D.\ candidate in Aerospace Science \& Engineering with the Space and Terrestrial Autonomous Robotic Systems (STARS) Laboratory at the University of Toronto Institute for Aerospace Studies, Toronto, Canada. His research focuses on uncertainty-aware global robotic mobility planning in planetary environments. Since 2017, he has been a visiting research student at NASA's Jet Propulsion Laboratory four times, working with the Robotic Mobility group. He earned his B.Eng.\ in Mechanical Engineering (major) and Geology (minor) from McGill University, Montreal, Canada.
\end{biographywithpic} 

\begin{biographywithpic}
{Shantanu Malhotra}{figs/shan_malhotra.jpg}
is a Principal Engineer at NASA’s Jet Propulsion Lab. He focuses on building mission critical ground data systems. For the past 24 years he has worked as a System Engineer on DSN’s Service Management System – which is responsible for prediction and configuration of the systems as part of realtime tracking. In the last 14 years he has worked as the System Engineer for the Treks---a series of Graphical Information Systems that support planetary science, mission design and public outreach for the Moon, Mars, Vesta, Titan and even Earth.
\end{biographywithpic} 

\begin{biographywithpic}
{Jonathan Kelly}{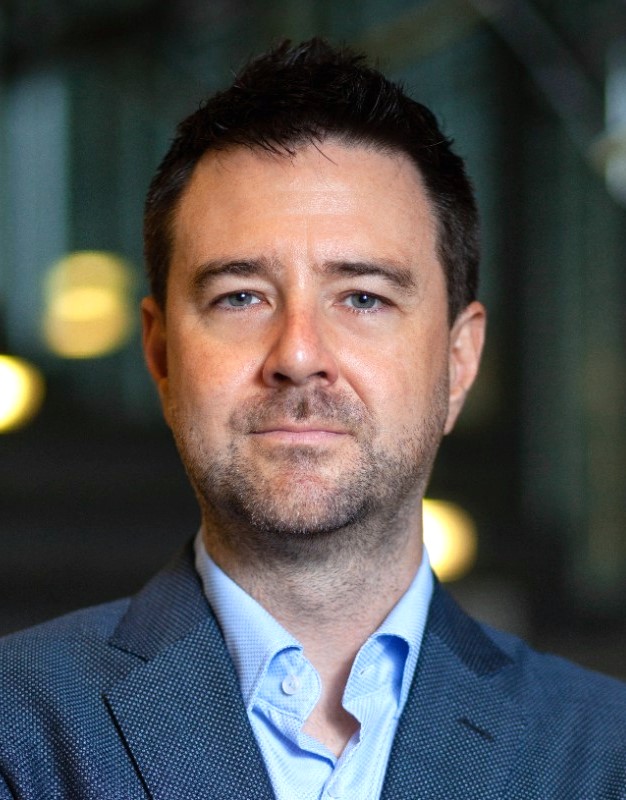}
received the Ph.D.\ degree in Computer Science from the University of Southern California, Los Angeles, USA, in 2011. From 2011 to 2013 he was a postdoctoral associate in the Computer Science and Artificial Intelligence Laboratory at the Massachusetts Institute of Technology, Cambridge, USA. He is currently an associate professor and director of the Space and Terrestrial Autonomous Robotic Systems (STARS) Laboratory at the University of Toronto Institute for Aerospace Studies, Toronto, Canada. He has held positions as a visiting research scientist at NASA’s Jet Propulsion Laboratory and as a software engineer at the Canadian Space Agency, Saint-Hubert, Canada.
\end{biographywithpic}

\end{document}